\documentclass[lettersize,journal]{IEEEtran}
\usepackage{amsmath,amsfonts}
\usepackage{algorithmic}
\usepackage{algorithm}
\usepackage{array}
\usepackage[caption=false,font=normalsize,labelfont=sf,textfont=sf]{subfig}
\usepackage{textcomp}
\usepackage{stfloats}
\usepackage{url}
\usepackage{verbatim}
\usepackage{graphicx}

\usepackage{import}
\usepackage[detect-all, separate-uncertainty=true, binary-units]{siunitx}
\usepackage{amsmath}
\usepackage{mathtools}
\usepackage{mleftright}
\usepackage{commath}
\usepackage{amssymb}
\usepackage{circuitikz}
\usepackage{hyperref}
\hypersetup{
    colorlinks=false,
    pdfborder={0 0 0},
    pdfborderstyle={/S/U/W 0},
}
\usepackage[]{cleveref}
\Crefname{listing}{listing}{listings}
\Crefname{figurepanel}{Fig.}{Fig.}
\crefname{figurepanel}{fig.}{fig.}
\usepackage{tabularx}
\usepackage{afterpage}
\usepackage{glossaries}

\usepackage[textsize=footnotesize]{todonotes}
\usepackage[backend=bibtex,style=ieee,natbib=true,maxbibnames=3]{biblatex}
\newacronym{isi}{ISI}{inter-spike interval}
\newacronym{adex}{AdEx}{adaptive exponential leaky integrate-and-fire}
\newacronym{sram}{SRAM}{static random-access memory}
\newacronym{mosfet}{MOSFET}{metal-oxide-semiconductor field-effect transistor}
\newacronym{ota}{OTA}{operational transconductance amplifier}
\newacronym{psp}{PSP}{postsynaptic potential}
\newacronym{ann}{ANN}{artificial neural network}
\newacronym{lif}{LIF}{leaky integrate-and-fire}
\newacronym{asic}{ASIC}{application-specific integrated circuit}
\newacronym{cmos}{CMOS}{complementary metal-oxide-semiconductor}

\input{preamble/circuits.tex}
\input{preamble/wisker.tex}
\input{preamble/panels.tex}

\makeatletter
\def\do#1{\patchcmd{#1}{\thepage}{\null}{}{\GenericWarning{}{Could not patch \string#1}}}
\docsvlist{\@oddhead,\@evenhead,\ps@headings,\ps@IEEEtitlepagestyle,\ps@IEEEpeerreviewcoverpagestyle}
\makeatother

\usepackage{titlesec}
\titlespacing{\section}{0pt}{2.2ex plus .3ex minus .1ex}{\dimexpr1.6ex-4pt plus .1ex}
\titlespacing{\subsection}{0pt}{2.1ex plus .1ex minus .1ex}{\dimexpr1.5ex-4pt plus .1ex}

\begin{document}

\let\oldcaption\caption
\renewcommand\caption[1]{\vspace{-1.6em}\oldcaption{#1}\vspace{-1.0em}}

\title{An accurate and flexible analog emulation of AdEx neuron dynamics in silicon}

\newcommand{\correspondence}{\textsuperscript{$\star$}}
\newcommand{\affilKIP}{\textsuperscript{$\dagger$}}

\author{%
	Sebastian Billaudelle\affilKIP\correspondence,
	Johannes Weis\affilKIP,
	Philipp Dauer\affilKIP,
	Johannes Schemmel\affilKIP,~\IEEEmembership{Member,~IEEE}\\
	{\footnotesize%
	\correspondence \texttt{sebastian.billaudelle@kip.uni-heidelberg.de}
	\affilKIP Kirchhoff-Institute for Physics, Heidelberg University, Germany
	}

	\vspace{-0.7cm}
}

\maketitle

\begin{abstract}
Analog neuromorphic hardware promises fast brain emulation on the one hand and an efficient implementation of novel, brain-inspired computing paradigms on the other.
Bridging this spectrum requires flexibly configurable circuits with reliable and reproducible dynamics fostered by an accurate implementation of the targeted neuron and synapse models.
This manuscript presents the analog neuron circuits of the mixed-signal accelerated neuromorphic system BrainScaleS-2.
They are capable of flexibly and accurately emulating the \gls{adex} model equations in combination with both current- and conductance-based synapses, as demonstrated by precisely replicating a wide range of complex neuronal dynamics and firing patterns.
\end{abstract}


\section{Introduction}

\begin{figure*}[!b]
\centering
	\begin{tikzpicture}
		\node[panel,anchor=north west] (a) at (0, 0) {
			\includegraphics[width=5.2cm]{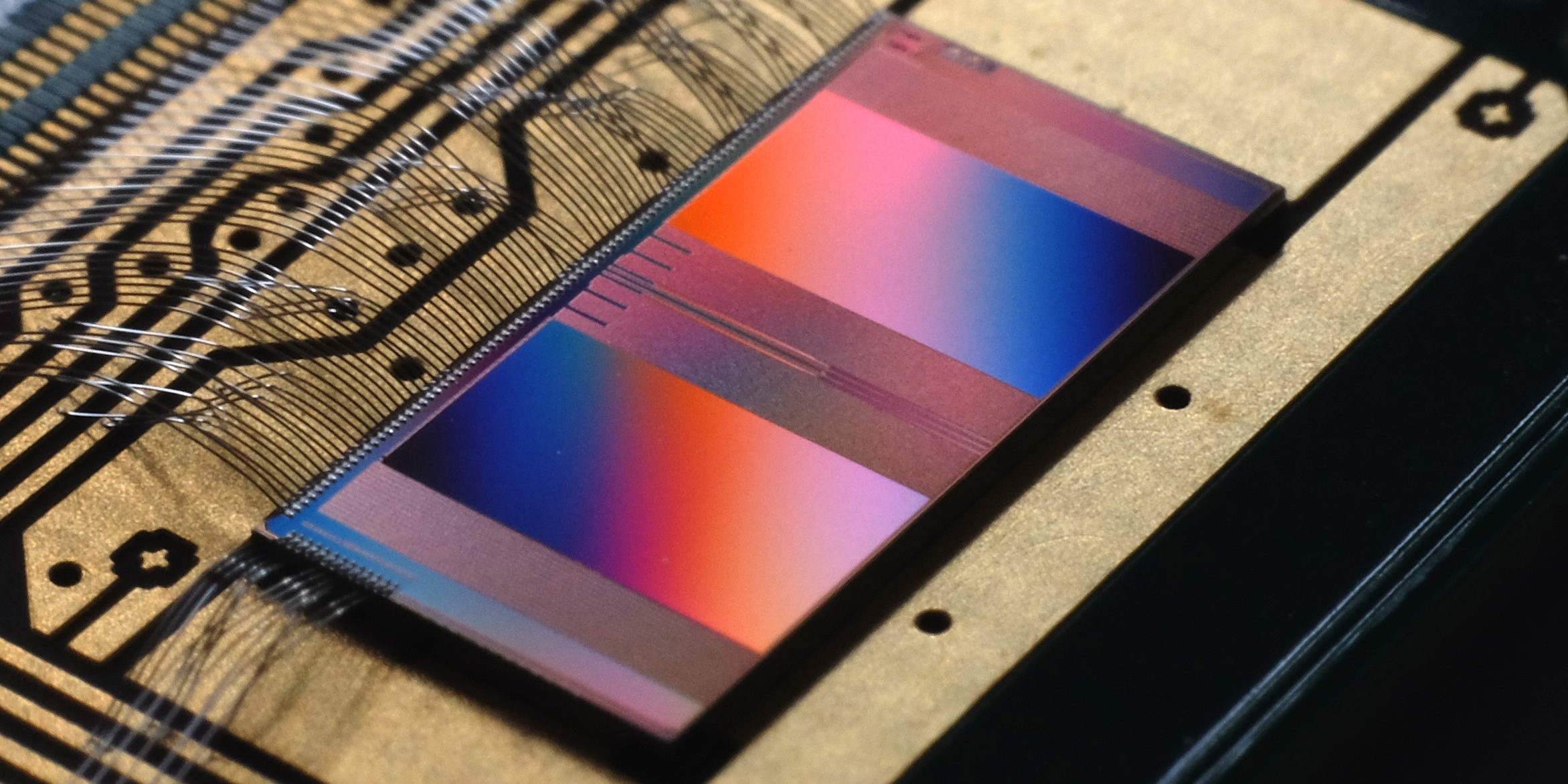}
		};
		\node[label,handle={fig:chip-photo},white,caption={
			Photograph of a BrainScaleS-2 \gls{asic}.
		}] at (a.north west) {A};
		
		\node[panel,anchor=north east] (c) at (\textwidth, 0) {
			\subimport*{figures/overview}{overview.tex}
		};

		\coordinate (tmp) at (0, 0);

		\node[panel,anchor=south west] (b) at (c.south west -| tmp) {
			\begin{tikzpicture}[
            font={\rmfamily\fontsize{8}{8}\selectfont},
	    transform shape,
            scale=0.72,
            >=stealth,
        ]

	\usetikzlibrary{positioning,fit,calc}

	\tikzset{block/.style={font={\scshape\fontsize{8}{8}\selectfont},align=center}}
	\tikzset{box/.style={draw=black!90}}
	\tikzset{box/.style={rectangle,line width=0.8pt,rounded corners=1pt,draw,inner sep=0pt}}

	\tikzset{block label/.style={fill=white,font={\rmfamily\footnotesize},inner sep=0.05cm}}
        \pgfdeclarelayer{background layer}
        \pgfsetlayers{background layer,main}

	\node[block,box,minimum width=3.0cm,minimum height=2.0cm] (syn) at (0,0) {};
	\node[block,minimum width=3.0cm,minimum height=0.4cm,below=0.2cm of syn] (nrn) {};
	\node[block,box,minimum width=3.0cm,minimum height=0.4cm,below=0.1cm of nrn] (cm) {parameter storage};

	\node[block,minimum width=0.5cm,minimum height=2.0cm,left=0.05cm of syn] (drv) {};
	\node[block,minimum width=0.2cm,minimum height=2.0cm,left=0.05cm of drv] (padi) {};
	
	\node[block,box,minimum width=3.0cm,minimum height=0.3cm,above=0.1cm of syn] (cadc) {cadc};
	\node[block,box,minimum width=3.0cm,minimum height=0.4cm,above=0.2cm of cadc] (ppu) {ppu};

	\node[anchor=north west,gray,rotate=90,yshift=-0.8ex] (anc) at (cm.south east) {\scshape analog network core};
	\node[block,box,gray,dashed,fit={(padi) (cadc) (anc)},inner sep=0.1cm] (anc bound) {};

	\node[block,box,anchor=south east,minimum width=1.7cm,minimum height=0.8cm] (event)
		at ($(nrn.south west -| anc bound.west) + (-0.2,0.2)$) {event \\ handling};
	\node[block,box,anchor=south east,minimum width=2.2cm,minimum height=0.4cm] (link)
		at ($(anc bound.south west) - (0.2,0.0)$) {I/O};
	\node[block,box,anchor=north east,minimum width=1.7cm,minimum height=1.2cm] (mem)
		at ($(anc bound.north west) - (0.2,0.0)$) {config.\\memory\\control.};

	\foreach \x in {0,1,...,5} {
		\draw[] (nrn.south west) ++ (\x*0.5,0.0) ++ (0.25,0.25) circle (0.2cm);
		\draw[gray,stealth-] (syn.south west) ++ (\x*0.5+0.25,-0.22) -- ++(0.0,2.22);
		
		\foreach \y in {0,1,...,3}
			\draw[gray] (syn.south west) ++ (\x*0.5+0.25,0.25+\y*0.5) circle (0.02cm);
	}

	\foreach \y in {0,1,...,3} {
		\draw[] (drv.south west) ++ (0.0,\y*0.5) ++ (0.1,0.05) -- ++(0.0,0.4) -- ++(0.3,-0.2) -- cycle;
		\draw[gray] (syn.south west) ++ (-0.1,\y*0.5+0.25) -- ++(3.0,0.0) -- ++(0.1,0.0);
	}

	\draw ($(event.east |- drv.south west) + (0.0,0.25)$) -- ($(drv.south west) + (-0.2,0.25)$) -- ++(0.0,1.5);
	\draw[,stealth-] (event.east |- nrn.west) -- (nrn.west);
	\draw[,stealth-stealth] (link.north -| event) -- (event);
	\foreach \y in {0,1,...,3}
		\draw[stealth-,line cap=rect] (drv.south west) ++ (0.1,\y*0.5+0.25) -- ++(-0.3,0.0);

	\draw[stealth-stealth] (event.west) -- ++(-0.20, 0.0) coordinate (x) -- (link.north -| x);
	\draw[-stealth] (x) -- (x |- ppu) -- (ppu);
	\draw[-stealth] (x) -- (x |- mem) -- (mem);
	\draw[-stealth] (mem.east) ++ (0.0, 0.1) -- ++(0.2,0.0);
	\draw[stealth-] (mem.east) ++ (0.0,-0.1) -- ++(0.2,0.0);

	\node[block label] at (syn) {\scshape\fontsize{8}{8}\selectfont synapse array};
	\node[block label] at (nrn) {\scshape\fontsize{8}{8}\selectfont neurons};
\end{tikzpicture}
		};
		\node[label,handle={fig:architecture},caption={
			Block-level schematic of the neuromorphic system.
		}] at (b.north west) {B};
		
		\node[label,handle={fig:block-level},caption={
			Block-level schematic of the silicon neuron.
		}] at (c.north west) {C};
	\end{tikzpicture}
\caption{\panelcaptions}
\label{fig_sim}
\end{figure*}

\IEEEPARstart{S}{piking} neuron models attempt to replicate the time-continuous dynamics and the event-based, asynchronous communication scheme of their archetype.
Building on the simpler \gls{lif} equation, the \gls{adex} model captures these fundamental properties and can reproduce many of the dynamics and firing patterns found in electrophysiological recordings of biological neurons \cite{naud2008firing}.
It describes the dynamics of the membrane potential $V$ as
\begin{align}
	C\od{V}{t} &= -g_\text{l} \left( V-E_\text{l} \right)
		\,+\, g_\text{l}\Delta_\text{T} \operatorname{e}^{\frac{V-V_\text{T}}{\Delta_\text{T}}}
		\,-\, w
		\,+\, I \,, \label{eq:adex-v}
\end{align}
with membrane capacitance $C$, leak potential $E_\text{l}$, leak conductance $g_\text{l}$, soft threshold $V_\text{T}$, and exponential slope $\Delta_\text{T}$.
The dynamics of the adaptation current $w$ are governed by
\begin{align}
	\tau_\text{w}\od{w}{t} &= a \left( V-E_l \right) - w \,, \label{eq:adex-w}
\end{align}
with time constant $\tau_\text{w}$ and subthreshold adaptation strength $a$.
The time-continuous dynamics are accompanied by spike-triggered jump conditions $V \rightarrow V_\text{r}$ and $w \rightarrow w + b$.

Synaptic interaction is typically modeled by weighting and accumulating the presynaptic spike trains of each synapse $i$, $S_i(t) = \sum_j \delta(t - t_j)$, and by convolving them with an interaction kernel, typically an exponential decay $ s(t) = \sum_i w_i S_i(t) \circledast \exp(t / \tau_\text{syn}) $.
This integrator trace is then translated into a current onto the membrane either directly (\emph{current-based}) or by modulating a conductance pulling the membrane towards a synaptic reversal potential (\emph{conductance-based}):
\begin{align}
	I_\text{cuba} = \hat I \cdot s(t)\,, \quad\text{or}\quad I_\text{coba} = \hat g \cdot s(t) \cdot \left[ E_\text{syn} - V \right] \,.\label{eq:synaptic-current}
\end{align}
The \gls{adex} equations and other complex neuron models have inspired a plethora of analog neuromorphic implementations \cite{wijekoon2008integrated,folowosele2009switched,van2010log,chicca2014neuromorphic,aamir2018mixed} and also lay the foundation for the silicon neuron presented in this manuscript.
The discussed silicon neuron can accurately reproduce the original model dynamics and can be configured for a wide range of parameterizations.

The neuron circuits represent one of the central components of the analog core of BrainScaleS-2 (\Cref{fig:chip-photo}), a mixed-signal neuromorphic system emulating neuronal and synaptic dynamics on 1000-fold accelerated time scales compared to the biological nervous system.
The silicon neurons are embedded into the rich infrastructure provided by the neuromorphic \gls{asic} (\Cref{fig:architecture}).
They receive weighted stimuli via the synapse array and relay their output spikes to the chip-internal event routing fabric.
While partially controlled and interfaced through full-custom digital backend logic (\Cref{fig:block-level}), the neuronal dynamics are realized by analog circuits.

\section{Silicon Implementation}

\noindent
The design can be separated into multiple functional blocks mainly corresponding to the individual terms of the model equations.
The circuit is implemented in a \SI{65}{\nano\meter} bulk \gls{cmos} technology and occupies an active silicon area of approximately \SI{2575}{\square\micro\meter}.

\subsection{Bulk-driven OTA as a building block}

\noindent
The mostly linear nature of the differential equations behind the model dynamics emphasizes the need for an area- and energy-efficient, widely linear component, mainly to translate voltages into proportional currents.
For that purpose, preceding silicon neurons have often already relied on \glspl{ota} \cite{millner2010vlsi,aamir2018mixed}.
Saturation effects and the limited linear range of standard \gls{ota} designs, however, can dramatically distort the underlying dynamics when compared to the original model equations.
The presented circuits address this issue by relying on a bulk-driven differential pair (M1, M2) to linearize the characteristics of the \gls{ota} (\Cref{fig:ota-schematic}).
Bulk-driven circuits exploit the body effect to modulate a transistor's drain current \cite{khateb2010utilizing}.
In the present case, we make use of the reduced transconductance $g_\text{mb} \ll g_\text{m}$ to increase the linear input range of the \gls{ota} (\Cref{fig:ota-transconductance}), lifting the requirement for more complex and less area- or energy-efficient linearization techniques but requiring careful biasing and verification to always maintain reverse biased body diodes.
Depending on the exact, application-specific dimensioning, the discussed circuit exhibits a footprint down to less than $\SI[parse-numbers=false]{10\times10}{\square\micro\meter}$, despite the requirement for isolated and thus spatially separated n-wells for the two transistors of the differential pair.
It serves with only slight modifications to the specific instantiation as a common building block for the silicon neuron, e.g. realizing the leak conductance of the membrane dynamics (\Cref{eq:adex-v}).

\setcounter{panel}{0}
\begin{figure}[!t]
	\centering
	\begin{tikzpicture}
		\node[panel,anchor=north west] (a) at (0, 0) {
			\scalebox{0.75}{\begin{tikzpicture}[scale=0.75, transform shape]
	\draw (-1.3, 0) node[rground] {} to ++(0, 0.0)
		to[Tnmos, n=m3] ++(0, 1.5)
		to ++(0.0, 0.2)
		to[Tpmos, mirror, bulk, n=m1] ++(0, 1.5)
		(m1.north east) node[left] {M1};
	\draw ( 1.3, 0) node[rground] {} to ++(0, 0.0)
		to[Tnmos, mirror, n=m4] ++(0, 1.5)
		to ++(0.0, 0.2)
		to[Tpmos, bulk, n=m2] ++(0, 1.5)
		(m2.north east) node[right] {M2};
	\draw (m1.G) to[short] ($(m1)!0.5!(m2)$) coordinate (midpoint) to (m2.G);
	\draw (midpoint) to[short, *-] (0, 0) node[rground] {};

	\draw (m1.bulk) to[short, -o] ++(-0.5, 0) node[left] {$-$};
	\draw (m2.bulk) to[short, -o] ++( 0.5, 0) node[right] {$+$};

	\draw (m1.S) to ++(0, 0.2) to ($(m1.S)!0.5!(m2.S) + (0, 0.2)$);
	\draw (m2.S) to ++(0, 0.2) to ($(m1.S)!0.5!(m2.S) + (0, 0.2)$) to[short, *-o] ++(0, 0.5) node[above] {$I_\text{b}$};

	\draw (-3.3, 0) node[rground] {} to ++(0, 0.0)
		to[Tnmos, mirror, n=m5] ++(0, 1.5)
		to ++(0, 2.4)
		to[Tpmos, mirror, n=m7] ++(0, 1.5)
		to ++(0, 0.2)
		edge[thick,line cap=round] ++(-0.15, -0.2)
		edge[thick,line cap=round] ++( 0.15, -0.2);
	\draw (m3.D) to[short, *-] (m3.D -| m3.G) to[short, -*] (m3.G) to (m5.G);

	\draw ( 3.3, 0) node[rground] {} to ++(0, 0.0)
		to[Tnmos, n=m6] ++(0, 1.5)
		to ++(0, 2.4)
		to[Tpmos, n=m8] ++(0, 1.5)
		to ++(0, 0.2)
		edge[thick,line cap=round] ++(-0.15, -0.2)
		edge[thick,line cap=round] ++( 0.15, -0.2);
	\draw (m4.D) to[short, *-] (m4.D -| m4.G) to[short, -*] (m4.G) to (m6.G);

	\draw (m7.D) to[short, *-] (m7.D -| m7.G) to[short, -*] (m7.G) to (m8.G);

	\draw ($(m8.D)!0.5!(m6.D)$) to[short, *-o] ++(0.6, 0) node[right] {$I_\text{out}$};
\end{tikzpicture}}
		};
		\node[label,xshift=0.3cm,handle={fig:ota-schematic},caption={
			Bulk-driven \gls{ota} serving as a building block for the silicon neuron (simplified schematic).
		}] at (a.north west) {A};

		\node[panel,anchor=north east] (b) at (\columnwidth, 0) {
			\includegraphics{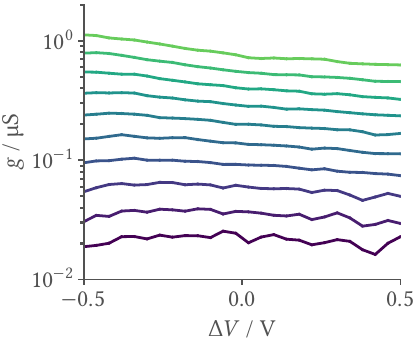}
		};
		\node[anchor=north east, inner sep=0pt] at (b.north east) {\includegraphics{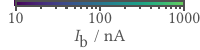}};
		\node[label,handle={fig:ota-transconductance},caption={
			Small signal transconductance measured for different bias currents.
		}] at (b.north west) {B};
	\end{tikzpicture}
	\caption{\panelcaptions}
	\label{fig_sim}
\end{figure}

\subsection{Adaptation}

\noindent
The adaptation state variable $w$ (\Cref{eq:adex-w}) also resembles leaky integrator dynamics.
While taking the form of a current in the model equations, it is in our silicon neuron represented as a voltage $V_\text{w}$.
The core dynamics are, hence, implemented through a low-pass filter, directly relying on a capacitor and the bulk-driven \gls{ota} as a pseudo-conductance (\Cref{fig:adaptation-schematic}), allowing to tune the adaptation time constant through its bias (\Cref{fig:adaptation-tau}).
We fitted OTA2 with a second, approximately twelve-fold stronger output stage to repurpose the circuit to also generate the adaptation current $I_\text{w} = g_\text{w}\mleft(I_\text{b}^\tau\mright) \left( V_\text{ref} - V_\text{w} \right)$ directed onto the membrane.

The adaptation state is driven by the membrane-dependent subthreshold contributions on one hand and spike-triggered increments on the other.
The former are realized through OTA1, whereby the coupling strength $g_\text{a}\mleft(I_\text{b}^\text{a}\mright)$ is controlled via the respective bias current (\Cref{fig:adaptation-subthreshold}).
The time-continuous dynamics of the adaptation current hence result as
\begin{align}
	\tau_\text{w} \od{I_\text{w}}{t} &= - \tau_\text{w} g_\text{w} \od{V_\text{w}}{t} = \pm \underbrace{g_\text{a} \frac{g_\text{w}}{g_\tau}}_{\equiv a} \cdot \left( V_\text{m} - E_\text{l}^\text{adapt} \right) - I_\text{w} \,,
\end{align}
with $ \tau_\text{w}\mleft(I_\text{b}^\tau\mright) \equiv C_\text{w} / g_\tau\mleft(I_\text{b}^\tau\mright) $.
These dynamics are augmented by spike-triggered in- or decrements of $V_\text{w}$ resulting from short current pulses $I_\text{b}$ with configurable width and amplitude.

\begin{figure}[!t]
	\centering
	\begin{tikzpicture}
		\node[panel,anchor=north west] (a) at (0, 0) {
			\scalebox{0.75}{
				\begin{tikzpicture}[scale=0.75, transform shape]
	\node[gm amp] (ota1) at (0,0) {OTA1};

	\draw (ota1.-) to ++(-0.5, 0) node[cute spdt up, anchor=in, scale=-1, circuitikz/bipoles/cuteswitch/thickness=0.3, circuitikz/bipoles/cuteswitch/height=0.3] (s1) {};
	\draw (ota1.+) to ++(-0.5, 0) node[cute spdt up, anchor=in, scale=-1, circuitikz/bipoles/cuteswitch/thickness=0.3, circuitikz/bipoles/cuteswitch/height=0.3] (s2) {};
	
	\draw[stealth-] (s1) ++(0, 0.2) -- ++(0, 0.8) node[right, rotate=90] {\texttt{invert\_a}};
	\draw[dashed] (s1.mid) -- (s2.mid);

	\draw (s1.out 1) ++(-1.5, 0) coordinate (vl) node[left] {$E_\text{l}^\text{adapt}$};
	\draw (s2.out 2) ++(-1.5, 0) coordinate (vm) node[left] {$V_\text{m}$};

	\draw (s1.out 1) to ++(-0.2, 0) coordinate (tmp) to (tmp |- s2.out 2) to (s2.out 2);
	\draw (vm) to[short, o-*] (vm -| tmp);

	\draw (s1.out 2) to ++(-0.4, 0) coordinate (tmp) to (tmp |- s2.out 1) to (s2.out 1);
	\draw (vl) to[short, o-*] (vl -| tmp);

	\draw (ota1.out) to[short, label=$V_\text{w}$] ++(5.3, 0) node[double gm amp, anchor=-] (ota2) {OTA2};
	\draw (ota1.out) ++(3, 0) to[C, l=$C_\text{w}$, *-] ++(0, -1.5) node[rground] {};
	\draw (ota2.out2) to[short, -o] ++(1, 0) node[right] {$V_\text{m}$};
	\draw (ota2.out1) to ++(0, +1.6) coordinate (fb) to (fb -| ota2.-) to[short, -*] (ota2.-);
	\draw (ota2.out2) ++(-0.9, 0) node[right] {\footnotesize $g_\text{w}$};
	\draw (ota2.out1) ++(-0.9, 0) node[right] {\footnotesize $g_\tau$};
	\draw (ota2.+) to[short, -*] ++(-0.3, 0) node[left] {$V_\text{ref}$};

	\draw (ota1) ++(0.0, 0.72) to[short,-o] ++(0.0, 0.5) node[above] {$I_\text{b}^\text{a}$};
	\draw (ota2) ++(0.0, 0.72) to[short,-o] ++(0.0, 0.5) node[above] {$I_\text{b}^\tau$};

	\draw[-stealth] ($(ota2.out2) + (0.05, -0.2)$) -- ++(0.5, 0.0) node[pos=0.5,below] {\fontsize{8}{8}\selectfont $I_\text{w}$};
	\draw[stealth-] ($(ota2.out1) + (0.2, 0.6)$) -- ++(0.0, -0.5) node[pos=0.5,right] {\fontsize{8}{8}\selectfont $I_\tau$};

	\node[minimum width=2.8cm, minimum height=1.6cm, draw=black, thick] (spike triggered) at (3, 2.7) {};

	\draw (spike triggered) ++ (-0.1,-0.3) -- ++(0.4,0) -- ++(0,0.6) -- ++(0.4,0) -- ++(0, -0.6) -- ++(0.4, 0);
	\draw (spike triggered.90) node[below] {$I_\text{b}$} to[short,-o] ++(0, 0.5) node[above] {$I_\text{b}^\text{b}$};
	\draw (spike triggered.167) node[right] {\footnotesize\vphantom{gl}\texttt{pulse}} to[short,-o] ++(-0.7, 0);
	\draw (spike triggered.193) node[right] {\footnotesize\vphantom{gl}\texttt{invert}} to[short,-o] ++(-0.7, 0);
	\draw (spike triggered.270) to[short,-*] (spike triggered |- ota2.-);
	
	\draw[-stealth] ($(spike triggered.270) + (0.2, -0.2)$) -- ++(0.0, -0.5) node[pos=0.5,right] {\fontsize{8}{8}\selectfont $I_\text{b}$};
\end{tikzpicture}
			}
		};
		\node[label,handle={fig:adaptation-schematic},caption={
			Schematic of the adaptation circuit.
		}] at (a.north west) {A};

		\node[panel,anchor=north west] (b) at (0, -4.0) {
			\includegraphics{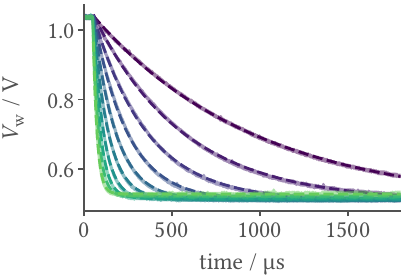}
		};
		\node[anchor=north east, inner sep=0pt] at (b.north east) {\includegraphics{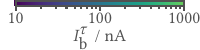}};
		\node[label,handle={fig:adaptation-tau},caption={
			Trajectory of the adaptation voltage after clamping and releasing from a fixed potential, measured for different bias currents.
		}] at (b.north west) {B};

		\node[panel,anchor=north east,minimum height=2cm] (c) at (\columnwidth, -4.0) {
			\includegraphics{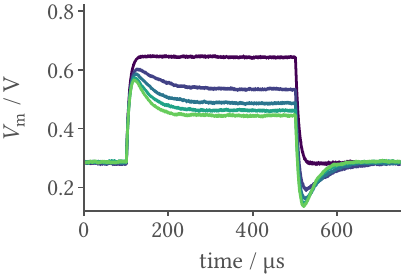}
		};
		\node[anchor=north east, inner sep=0pt] at (c.north east) {\includegraphics{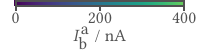}};
		\node[label,handle={fig:adaptation-subthreshold},caption={
			Transient response of a membrane to a step current measured for different subthreshold adaptation strengths.
		}] at (c.north west |- b.north) {C};
	\end{tikzpicture}
	\caption{\panelcaptions}
	\label{fig_sim}
\end{figure}

\subsection{Exponential feedback current}

\noindent
The exponential term is in its core implemented through the characteristics of a single \gls{mosfet} biased in weak inversion (\Cref{fig:exponential-schematic}, M4).
Its gate-to-source voltage is derived from the membrane via an \gls{ota} and a subsequent current-to-voltage conversion implemented straight-forwardly through a long-channeled transistor (M3) biased to operate in its linear region and thus acting as a resistor with on-resistance $r_\text{conv}$.
This compact conversion circuit allows tuning the exponential current's onset and slope (\Cref{fig:exponential-onset}, \Cref{fig:exponential-slope}):
\begin{align}
	I_\text{exp} (V_\text{m}) &= I_0 \cdot \exp \left( \frac{ 8 \cdot g_\text{ota} \cdot \left[ V_\text{m} - V_\text{exp} \right] }{ n \cdot V_\text{T} / r_\text{conv} } \right) \,.
\end{align}
This current is mirrored onto the membrane and can be gated either to disable the whole circuit or to ``pause'' the strong exponential feedback during the neuron's refractory period.
The current mirror formed by M1 and M2 rectifies the output current of OTA1 and, hence, effectively powers down the second half of the circuit for membrane potentials significantly below $V_\text{T}$.

\begin{figure}[!t]
	\centering
	\begin{tikzpicture}
		\node[panel,anchor=north west] (a) at (0, 0) {
			\scalebox{0.75}{
				\begin{tikzpicture}[scale=0.75, transform shape]
	\node[gm amp] (ota1) at (0,0) {OTA1};
	\draw (ota1.-) to[short, -o] ++(-0.5, 0) node[left] {$V_\text{exp}$};
	\draw (ota1.+) to[short, -o] ++(-0.5, 0) node[left] {$V_\text{m}$};
	\draw (ota1.up) to[short, -o] ++(0, 1) node[above] {$I_\text{b}^\text{exp}$};

	\draw (ota1.out) to ++(0.5, 0) to ++(0, 1) to[Tpmos, n=m1, l=M1, mirror] ++(0, 1) to ++(0, 0.5) node[vdd] (vdd) {};
	\draw (vdd) ++(2.5, 0) node[vdd] {}
		to ++(0, -0.5) to[Tpmos, n=m2, l=M2, invert] ++(0, -1)
		to ++(0, -2.0)
		to ++(0, -1.0) to[Tnmos, n=m3, l=M3, invert] ++(0, -1) to ++(0, -0.5) node[rground] (gnd) {};
	\draw (m1.D) to[short, *-] (m1.D -| m1.G) to[short, -*] (m1.G) to (m2.G);

	\draw (m3.G) to[short] ++(-0.3, 0) to ++(0.0, 0.5) node[vdd] {}; %

	\draw (gnd) ++(2.5, 0) node[rground] {} to ++(0, 1.5) to[Tnmos, n=m4, l=M4] ++(0, 1)
	to ++(0, 2) to[Tpmos, n=m5, l=M5, mirror] ++(0, 1) to ++(0, 0.5) node[vdd] {};
	\draw (m4.G) to[short, -*] (m4.G -| m2.D);

	\draw (vdd -| m5) ++(4.0, 0) node[vdd] {}
		to ++(0, -0.5) to[Tpmos, n=m6, l=M6, invert] ++(0, -1)
		to ++(0, -0.5) to[Tpmos, n=m7, l=M7, invert] ++(0, -1)
		to ++(0, -0.5) to[Tpmos, n=m8, l=M8, invert] ++(0, -1)
		to[short, -o, f=$I_\text{exp}$] ++(0, -1) node[below] {$V_\text{m}$};
	\draw (m5.D) to[short, *-] (m5.D -| m5.G) to[short, -*] (m5.G) to (m6.G);
	
	\node[right,xshift=0ex,yshift=1.5ex] at (m1.north) {1\hspace{1em}:};
	\node[left,xshift=0ex,yshift=1.5ex] at (m2.north) {8};
	
	\draw (m7.G) to[short,-o] ++(-0.2, 0.0) node[left] {\ttfamily pause};
	\draw (m8.G) to[short,-o] ++(-0.2, 0.0) node[left] {\ttfamily enableb};
\end{tikzpicture}
			}
		};
		\node[label,handle={fig:exponential-schematic},caption={
			Exponential term.
		}] at (a.north west) {A};
		
		\node[panel,anchor=north west] (b) at (0, -4.2) {
			\includegraphics{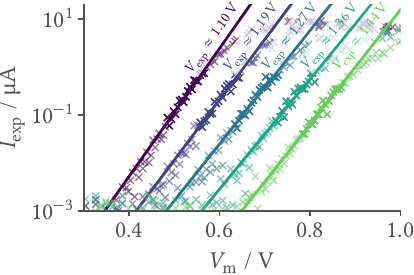}
		};
		\node[label,handle={fig:exponential-onset},caption={\hspace{-0.5ex}{\&}\hspace{-0.5ex}
		}] at (b.north west) {B};
		
		\node[panel,anchor=north east] (c) at (\columnwidth, -4.2) {
			\includegraphics{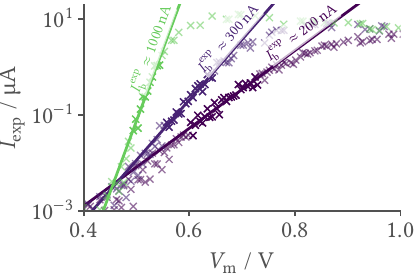}
		};
		\node[label,handle={fig:exponential-slope},caption={
			The exponential current is correctly reproduced across three orders of magnitude and can be parameterized via the reference potential and the OTA's bias.
			Saturation of the OTA, the current-to-voltage conversion, and the output stage limits $I_\text{exp}$ but -- due to the overall fast transients -- does not significantly impact spike timing.
		}] at (c.north west) {C};
	\end{tikzpicture}
	\caption{\panelcaptions}
	\label{fig:exponential}
\end{figure}

\subsection{Emulating current- and conductance-based synapses}

\noindent
The synaptic integrator circuits (\Cref{fig:synin-schematic}) represent the interface between the synapse array and the neuron's membrane circuits.
They integrate current pulses from their associated column of synapses \cite{friedmann2016demonstrating} and modulate the membrane in analogy to \Cref{eq:synaptic-current}.
Here, the low-pass filter directly exploits the synaptic line's capacitance, which can optionally be augmented with a dedicated capacitor.
The conductance (in \Cref{fig:synin-schematic} schematically drawn as a variable resistor) is realized through multiple series-connected p-channel \glspl{mosfet}.
They are biased in weak inversion to a constant gate-to-source voltage, similar to the predecessor circuit \cite{aamir2018mixed}.

OTA1 derives a current $I_\text{synin} = g_1 \cdot \Delta V_\text{syn}$ based on the deflection of the integrator voltage, directly implementing a \emph{current-based} output.
In that process, offsets can be compensated by tuning the voltage drop of the two source followers (M1, M2) via their bias currents.
\emph{Conductance-based} synapses (\Cref{fig:synin-coba}) are realized via an additional feedback path:
OTA2 can modulate the bias current of OTA1 based on the membrane potential such that\\[-2.0em]
\begin{align}
	g_1 &\propto g_2 \cdot \left( \smash[b]{\underbrace{\hat E_\text{syn} + I_\text{b}^\text{cuba} / g_2}_{\equiv E_\text{syn}}} - V_\text{m} \right)
		\vphantom{\left(\underbrace{\hat E}_{E}\right.} \,,
\end{align}\\[-0.5em]
where $g_1$ is assumed to be approximately proportional to the applied bias current.
Mixing the static bias current with an additional, modulated component allows to create a ``virtual'' reversal potential $E_\text{syn}$.
In the excitatory case, the latter usually extends far beyond the threshold voltage and thus falls outside the membrane's dynamic range.
Our design decouples the true zero crossing -- and with it the linear range -- of OTA2 from that normally never reached reversal potential.

\afterpage{
\begin{figure}[t]
\centering
	\begin{tikzpicture}
		\node[panel,anchor=north west] (a) at (0, 0) {
			\begin{tikzpicture}[scale=0.55, transform shape, anchor=center]
    \draw (-5, 4.0) node[vdd] {} to[short] ++(0, -1.5)
	to[Tnmos, mirror, n=m1, l=M1] ++(0, -1) to ++(0, -2.5) to[short, -o, f=$ $] ++(0, -1) node[below] {$I_\text{b}^\text{drop}$};
    \draw (-3, 4.0) node[vdd] {} to[short] ++(0, -1.5)
	to[Tnmos, mirror, n=m2, l=M2] ++(0, -1) to ++(0, -2.5) to[short, -o, f=$ $] ++(0, -1) node[below] {$I_\text{b}^\text{shift}$};
    \draw (m2.G) to (m2.G |- m2.S) to[short, -*] (m2.S);

    \draw (-7.0, 4.0) node[vdd] {} to[short] ++(0, -0.3) to[R, mirror, n=r1] ++(0, -1.2) to (r1 |- m1.G) to (m1.G);
    \draw (-8.5, 4.0) node[vdd] {} to[short] ++(0, -0.3) to[C, mirror, invert, n=c1] ++(0, -1.2) to (c1 |- m1.G) to[short, -*] (r1 |- m1.G);

	\draw (c1 |- m1.G) to[short, *-] ++(-0.5, 0) to[short, -o] ++(-1.0, 0) node[left] (vsyn label) {$V_\text{syn}$};

	\draw[-stealth] (vsyn label) ++(0.6, -0.3) -- ++(0.88, 0.0);
	\draw[] (vsyn label) ++(0.6, -0.5) -- ++(0.1, 0.0) -- ++(0.0, -0.4) -- ++(0.1, 0.0) -- ++(0.0, 0.4) -- ++(0.1, 0.0);
	\draw[] (vsyn label) ++(0.8, -0.75) node[right] {$I_\text{syn}$};
    
    \node[gm amp] (ota1) at (-0.3,0) {OTA1};
    \draw (ota1.-) to ++(-0.6, 0) node[above] {$\hat V_\text{syn}$} to[short, -*] (ota1.- -| m1.D);
    \draw (ota1.+) to ++(-0.6, 0) node[above] {$\hat V_\text{ref}$} to[short, -*] (ota1.+ -| m2.D);

    \node[gm amp, scale=-1] (ota2) at (2,2) {\scalebox{-1}[-1]{OTA2}};
    \draw (ota2.+) to[short, -o] ++(1.0, 0.0) node[right] {$\hat E_\text{syn}$};
	\draw (ota2.down) to[short, -o, f<=$ $] ++(0.0, 1.0) coordinate (biases) node[above] {$I_\text{b}^\text{coba}$};

    \draw (ota2.out) to (ota2.out -| ota1.up) coordinate (mix) to (ota1.up);
	\draw (mix) to[short, *-] (mix |- ota2.down) to[short, -o, f<=$ $] (mix |- biases)  node[above] {$I_\text{b}^\text{cuba}$};
    
    \draw (ota2.-) to[short, -*] (ota2.- |- ota1.out) coordinate (vm);
    \draw (ota1.out) to (vm) to[short, -o, f_=$I_\text{synin}$] ++(1.0, 0.0) node[right] {$V_\text{m}$};

	\node[wisker, wisker width=.6pt,wisker length=1.3cm] (w1) at (r1) {};
	\node[wisker, wisker width=.6pt,wisker length=1.3cm] (w2) at (c1) {};

	\draw[-stealth] ($(r1.left) + (0.5, 0.2)$) to[out=-18, in=18] ($(r1.right) + (0.5, -0.2)$);
	\node[inner sep=2pt, fill=white] at ($(r1) + (1.0, 0.0)$) {$\Delta V_\text{syn}$};
\end{tikzpicture}
		};
		\node[label,handle={fig:synin-schematic},caption={
			Simplified synaptic input circuit.
		}] at (a.north west) {A};
		
		\node[panel,anchor=north west] (b) at (0, -4.2) {
			\includegraphics{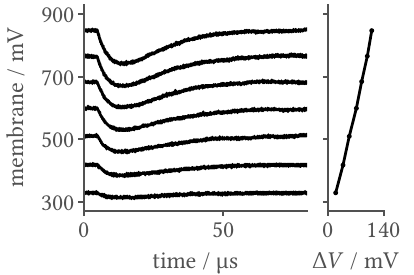}
		};
		\node[label,handle={fig:synin-coba},caption={
			Postsynaptic potentials measured for various membrane potentials, implementing conductance-based synaptic interaction.
		}] at (b.north west) {B};
		
		\node[panel,anchor=north east] (c) at (\columnwidth, -4.2) {
			\includegraphics{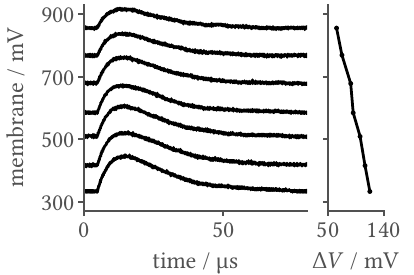}
		};
		\node[label,handle={fig:dummy},caption={
		}] at (c.north west) {};
	\end{tikzpicture}
	\caption{\panelcaptions}
	\label{fig:synin}
\end{figure}
}

\section{Results}

\begin{table}[b]
	\caption{Parameter ranges and variability across 128 neurons.}
	\label{tab:parameters}
	\begin{tabularx}{\columnwidth}{Xlllll}
		\hline
		parameter								& 		& {minimum}		& {maximum} 		& unit			\\
		\hline
		membrane time constant							& $\tau_\text{m}$	& \num{0.6 \pm 0.2} 	& \num{915 \pm 140} 	& \si{\micro\second}	\\
		leak potential								& $E_\text{l}$		& 0.0			& 1.0 			& \si{\volt}		\\ %
		firing threshold							& $V_\text{l}$		& 0.2			& 1.2 			& \si{\volt}		\\ %
		stimulus current							& $I_\text{stim}$	& 0.0			& \num{121 \pm 14} 	& \si{\nano\ampere}	\\
		synaptic time constant							& $\tau_\text{syn}$	& \num{0.29 \pm 0.03} 	& \num{538 \pm 98} 	& \si{\micro\second}	\\ %
		synaptic peak current	& $I_\text{syn}$	& \num{0.033 \pm 0.003}	& \num{1.15 \pm 0.03} 	& \si{\micro\ampere}	\\ %
		adaptation time constant							& $\tau_\text{w}$	& \num{22 \pm 3}	& \num{853 \pm 117} 	& \si{\micro\second}	\\ %
		subthreshold adaptation 						& $\abs{a}$			& \num{30 \pm 4}	& \num{1065 \pm 114}	& \si{\micro\siemens}	\\ %
		spike-triggered adapt.						& $\abs{b}$			& 0			& $\infty$ 		& \si{\micro\ampere}	\\ %
		exponential slope 							& $\Delta_\text{T}$	& \num{13\pm2}		& $>$\num{ 91\pm46}	& \si{\milli\volt}	\\ %
		\hline
	\end{tabularx}
\end{table}

\noindent
The neuron circuit's dynamics can be tuned via 16 analog bias currents and 8 voltages individually provided for each neuron instance and a set of digital controls (\SI{40}{\bit} of local \gls{sram}) to en- or disable parts of the circuit.
This large configuration space allows the flexible emulation of a wide range of model dynamics.
\Cref{tab:parameters} summarizes the parameter ranges achieved by the neuron design, measured across 128 neurons.
The effective parameter ranges can in many cases be further extended by selecting a reduced membrane capacitance, here configured for its maximum value of approximately \SI{2.47}{\pico\farad}.

With each neuron instance receiving an individual set of analog references generated by the on-chip parameter storage \cite{hock2013analog}, one may not only precisely tune the operating points to certain target dynamics but can also compensate for process corner effects as well as mismatch-induced fixed-pattern variability across neurons, a process we refer to as \emph{calibration}.
The flexible parameterization and calibration allows to configure the neuron circuits for a wide range of operating points.
In the following, we will demonstrate the reliability and accuracy of the silicon neuron and for that purpose, in each case, discuss the dynamics of a population of \num{128} calibrated neurons.

\subsection{LIF neuron dynamics}

\noindent
Calibrating the neuron circuits to a leak-over-threshold regime exhibits well-matching dynamics (\Cref{fig:leak-over-threshold-traces}) and narrow \gls{isi} distributions (\Cref{fig:leak-over-threshold-isis}).
The \glspl{isi} furthermore closely correspond to those predicted by evaluating the original model equations based on the calibration targets, here measured for membrane time constants spanning two full orders of magnitude (\Cref{fig:leak-over-threshold-sweep}).

A calibration of the synaptic input circuits, namely of $\tau_\text{syn}$, their effective amplitudes, and offsets, yields similarly well-matching postsynaptic currents.
This results in homogeneous \gls{psp} trajectories (\Cref{fig:psp-traces}), leaving the membrane's resting potential mostly unaffected (\Cref{fig:psp-baseline}) and yielding closely matching amplitudes (\Cref{fig:psp-amplitude}).

\begin{figure}[!t]
	\centering
	\begin{tikzpicture}
		\node[panel,anchor=north west] (a) at (0, 0) {
			\includegraphics[]{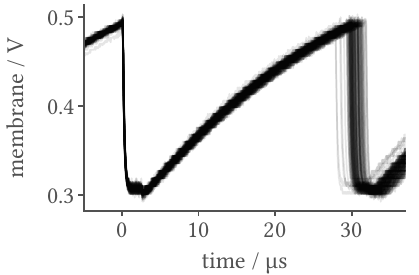}
		};
		\node[label,handle={fig:leak-over-threshold-traces},caption={
			Traces of 128 neurons calibrated for leak-over-threshold dynamics
		}] at (a.north west) {A};
		
		\node[panel,anchor=north west,yshift=-1ex] (b) at (a.south west) {
			\includegraphics[]{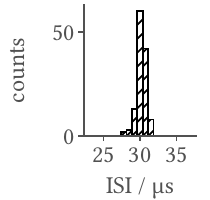}
		};
		\node[label,handle={fig:leak-over-threshold-isis},caption={
			and their \glspl{isi}.
		}] at (b.north west) {B};
		
		\node[panel,anchor=north east,yshift=-1ex] (c) at (a.south east) {
			\includegraphics[]{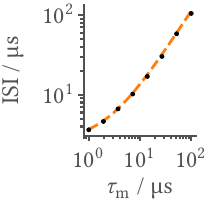}
		};
		\node[label,handle={fig:leak-over-threshold-sweep},caption={
			\Glspl{isi} for different choices of $\tau_\text{m}$ and comparison to predicted values.
		}] at (c.north west) {C};

		\node[panel,anchor=north east] (d) at (\columnwidth, 0) {
			\includegraphics[]{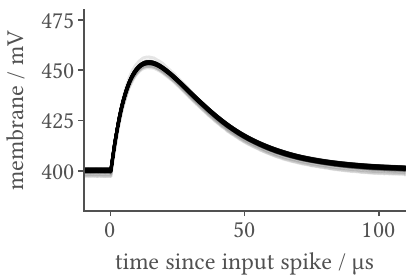}
		};
		\node[label,handle={fig:psp-traces},caption={
			\Gls{psp} traces of 128 neurons and distributions of
		}] at (d.north west) {D};

		\node[panel,anchor=north west,yshift=-1ex] (e) at (d.south west) {
			\includegraphics[]{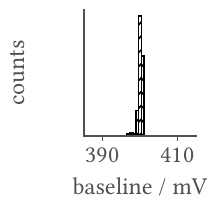}
		};
		\node[label,handle={fig:psp-baseline},caption={
			their baselines
		}] at (e.north west) {E};

		\node[panel,anchor=north east,yshift=-1ex] (f) at (d.south east) {
			\includegraphics[]{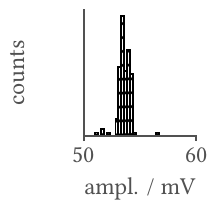}
		};
		\node[label,handle={fig:psp-amplitude},caption={
			and amplitudes.
		}] at (f.north west) {F};
	\end{tikzpicture}
	\caption{\panelcaptions}
	\label{fig_sim}
\end{figure}

\subsection{AdEx firing patterns}

\begin{figure}[!t]
	\centering
	\begin{tikzpicture}[y=1.7cm]
		\node[panel,anchor=north west] (a) at (0, 0) {
			\includegraphics[width=\columnwidth]{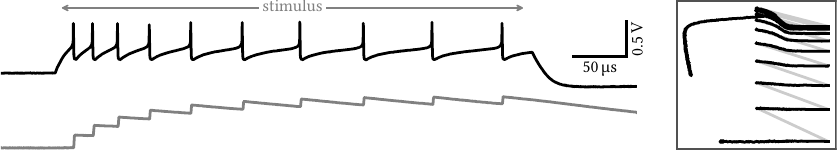}
		};
		\node[label,handle={fig:background-neuron-cajal},caption={
			--
		}] at (a.north west) {A};
		
		\node[panel,anchor=north west] (b) at (0, -1.0) {
			\includegraphics[width=\columnwidth]{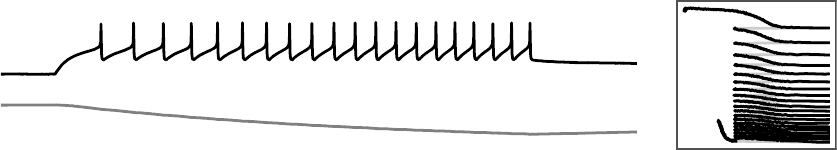}
		};
		\node[label,handle={fig:background-neuron-cajal},caption={
		}] at (b.north west) {B};
		
		\node[panel,anchor=north west] (c) at (0, -2.0) {
			\includegraphics[width=\columnwidth]{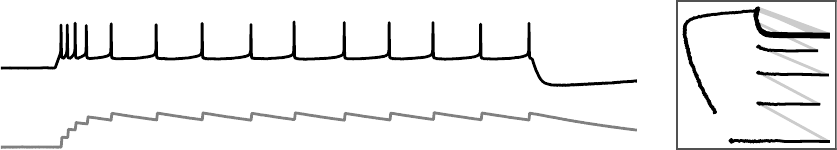}
		};
		\node[label,handle={fig:background-neuron-cajal},caption={
		}] at (c.north west) {C};
		
		\node[panel,anchor=north west] (d) at (0, -3.0) {
			\includegraphics[width=\columnwidth]{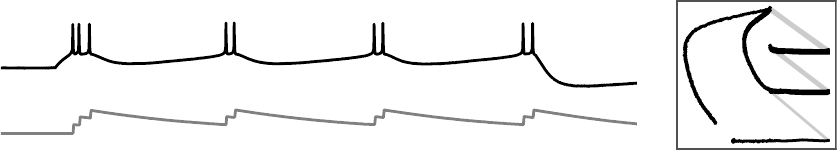}
		};
		\node[label,handle={fig:background-neuron-cajal},caption={
		}] at (d.north west) {D};

		\node[panel,anchor=north west] (e) at (0, -4.0) {
			\includegraphics[width=\columnwidth]{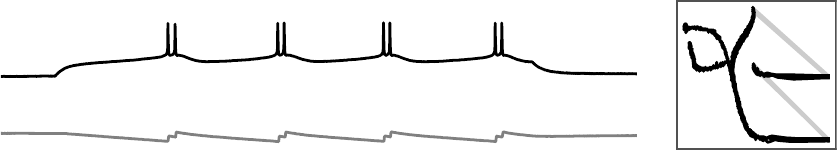}
		};
		\node[label,handle={fig:background-neuron-cajal},caption={
		}] at (e.north west) {E};

		\node[panel,anchor=north west] (f) at (0, -5.0) {
			\includegraphics[width=\columnwidth]{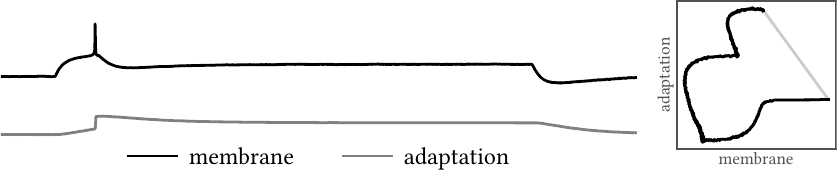}
		};
		\node[label,handle={fig:background-neuron-cajal},caption={
			\emph{Adaptation},
			\emph{delayed accelerating},
			\emph{initial burst},
			\emph{regular bursting},
			\emph{delayed regular bursting}, and
			\emph{transient spiking}.
		}] at (f.north west) {F};

		\draw[opacity=0.5, densely dashed, gray!80!white] (0.58, 0.0) -- ++(0.0, -5.94);
		\draw[opacity=0.5, densely dashed, gray!80!white] (5.646, 0.0) -- ++(0.0, -5.94);

	\end{tikzpicture}
	\caption{Membrane and adaptation state recordings as well as the reconstructed phase plane trajectory measured for a silicon neuron parameterized for various \gls{adex} firing patterns. \panelcaptions}
	\label{fig:adex-patterns}
\end{figure}

\noindent
We benchmarked the full \gls{adex} circuits by seeking to reproduce the different firing patterns discussed by \citeauthor{naud2008firing} \cite{naud2008firing}.
With only few exceptions, we relied on the originally published parameter sets and employed fully automated calibration routines to find suitable circuit configurations corresponding to those target dynamics.
\Cref{fig:adex-patterns} shows measured membrane and adaptation state traces for the first of 128 neurons.
These rich dynamics all emerge as a response to a constant step current and differentiate themselves only in the neurons' parameterizations.
All calibrated 128 neurons could -- with only slight deviations in the exact spike times and counts -- reproduce the desired firing patterns.

\section{Discussion}

\noindent
The circuits discussed in this manuscript faithfully emulate the \gls{adex} model equations and synaptic interaction.
They can be precisely calibrated to a wide range of operating points and capture the model dynamics with unprecedented accuracy.
Being part of the mixed-signal accelerated neuromorphic system BrainScaleS-2, they have proven themselves in a wide range of experimental studies ranging from the emulation of structured neurons to the machine-learning inspired training of spiking neural networks \cite{billaudelle2020versatile,pehle2022brainscales}.

\section*{Acknowledgements}
\footnotesize
\noindent
We thank
A. Grübl,
G. Kiene,
Y. Stradmann,
the Electronic Vision(s) group,
and all others involved
for their contributions.

This work has received funding from the European Union research and innovation funding
H2020 945539 (HBP SGA3),
DFG project EXC 2181/1 – 390900948 (STRUCTURES),
the Helmholtz Association project SO-092 (ACA),
and the Manfred Stärk Foundation.

\AtNextBibliography{\footnotesize}
\printbibliography

\vfill

\end{document}